\documentclass{article}
\usepackage{ijcai24}

\usepackage{times}
\usepackage{soul}
\usepackage{url}
\usepackage[hidelinks]{hyperref}
\usepackage[utf8]{inputenc}
\usepackage[small]{caption}
\usepackage{graphicx}
\usepackage{amsmath}
\usepackage{amsthm}
\usepackage{booktabs}
\usepackage{algorithm}
\usepackage{algorithmic}
\usepackage[switch]{lineno}
\usepackage{color}
\usepackage{natbib}
\usepackage{multirow}

\title{A Survey on Long Video Generation: Challenges, Methods, and Prospects}

\author{
Chengxuan Li$^{1,2,3}$
\and
Di Huang$^{5}$\and
Zeyu Lu$^{4,6}$\and
Yang Xiao$^{3,7}$\and
Qingqi Pei$^{1,2,3 *}$ \And
Lei Bai$^{4}$\thanks{Equal corresponding authors}\\
\affiliations
$^1$State Key Lab of Integrated Service Networks, Xian, China\\
$^2$Shannxi Key Laboratory of Blockchain and Secure Computing, Xian, China\\
$^3$Xidian University, Xian, China\\
$^4$Shanghai AI Laboratory, Shanghai, China\\
$^5$The University of Sydney, Sydney, Australia\\
$^6$Shanghai Jiao Tong University, Shanghai, China\\
$^7$The Engineerin Research Center of Trusted Digital Economy, Xidian, China\\
\emails
Emails: qqpei@mail.xidian.edu.cn,
baisanshi@gmail.com
}

\begin{document}

\maketitle

\begin{abstract}
Video generation is a rapidly advancing research area, garnering significant attention due to its broad range of applications.
One critical aspect of this field is the generation of long-duration videos, which presents unique challenges and opportunities.
This paper presents the first survey of recent advancements in long video generation 
and summarises them into two key paradigms: 
\textit{divide and conquer} or \textit{temporal autoregressive}.

We delve into the common models employed in each paradigm, including aspects of network design and conditioning techniques. Furthermore, we offer a comprehensive overview and classification of the datasets and evaluation metrics which are crucial for advancing long video generation research. 
Concluding with a summary of existing studies, we also discuss the emerging challenges and future directions in this dynamic field. 
We hope that this survey will serve as an essential reference for researchers and practitioners in the realm of long video generation.
\end{abstract}
\section{Introduction}
The realm of computer vision and artificial intelligence has experienced transformative growth, particularly in the area of video generation. Recently, there has been a surge in the development of algorithms that are capable of producing high-quality and realistic video sequences. 
Notably, the generation of long videos, characterized by their extended duration and complex content, has posed new challenges and inspired novel research directions in the community.

Nonetheless, there are still gaps in the research on long video generation. One of the gaps in current research is the absence of a standard definition for \textit{long videos}. The distinction between long and short videos often relies on relative measures in different works, such as frame count (e.g., 512, 1024, or 3376 frames) or duration (e.g., 3, 5 minutes), compared to shorter videos (e.g., 30, 48, or 64 frames).

\begin{figure}[t]
    \centering
    \includegraphics[scale=0.46]{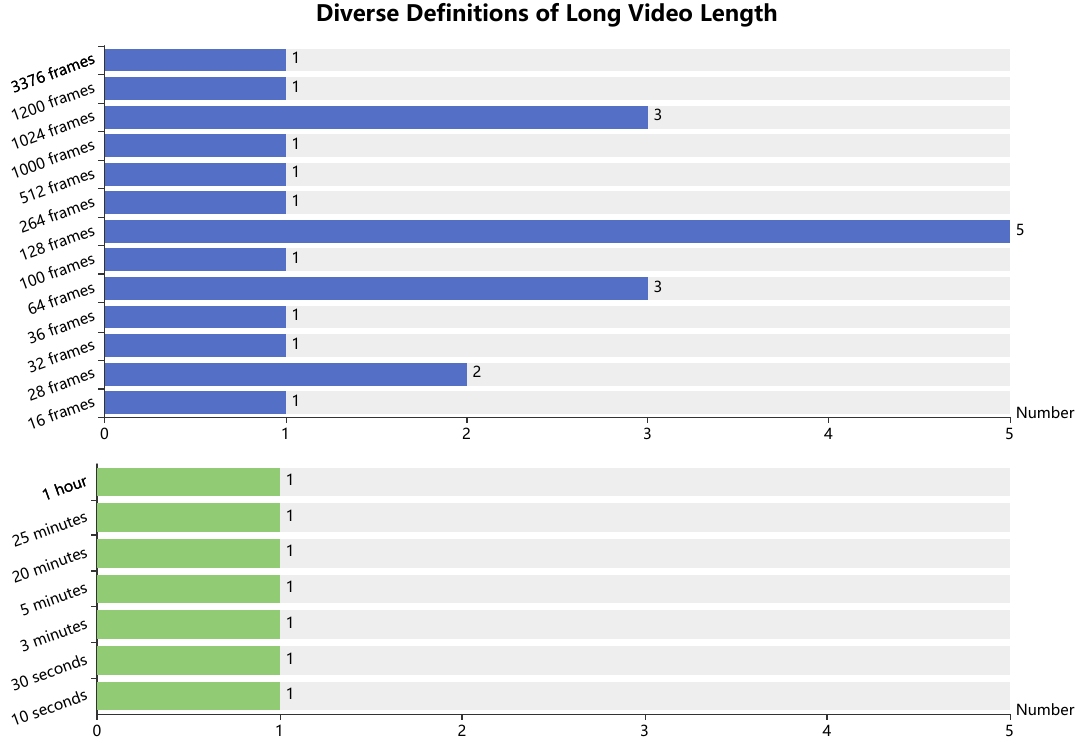}
    \caption{
    Overview of Long Video Length Definitions in Research. This figure compiles definitions of long video lengths from 51 studies, showing diverse standards. Out of these, 29 studies provide specific lengths: 7 in terms of duration and 22 in frame counts. The remaining studies do not specify video lengths.
    }
    \label{figure1}
\end{figure}

Considering the diversity in research standards, we summarize the length of the videos in existing researches that have claimed long video generation in Figure 1, based on which we propose a definition of long videos. Specifically, videos are classified as 'long' if their duration exceeds 10 seconds, assuming a standard frame rate of 10fps, or equivalently, if the video comprises more than 100 frames. This definition aims to provide a clear benchmark for the identification of long videos in various research contexts.

Under this definition, significant progress has been made for long video lengths. \cite{NUWA-XL} proposed a divide-and-conquer diffusion structure, specifically trained on long videos to eliminate the gap between inference and training, successfully generating videos with up to 1024 frames.
\cite{vlogger} leverage the strong capabilities of Large Language Models (LLM), extending input text into scripts to guide the generation of minute-level long videos. 
Recently, Sora \citep{Sora} achieves high-fidelity and seamlessly generates long videos up to one minute in duration, featuring high-quality effects such as multi-resolution and shot transitions.
Additionally, numerous outstanding research has introduced new structures and ideas on existing video generation models, paving the way for long video generation.

Even so, the generation of long videos still faces numerous challenges. At its core, the inherent multidimensional complexity of long videos imposes substantial demands on hardware resources for processing and generation, leading to a significant increase in training and generation costs in time and resources. This raises the challenge of generating long videos within the constraints of existing resources. Furthermore, the scarcity of long video datasets fails to meet training requirements, preventing researchers from directly obtaining optimal parameters to support the generation of long video models. In this scenario, when the generated video length exceeds certain thresholds, it is hard to keep the temporal consistency, continuity, and diversity for the long video generation. Moreover, current research has surfaced several phenomena that deviate from established physical laws of the real world, posing unforeseen challenges not yet comprehended or directly manipulable by existing methodologies.
Therefore, research on long video generation is still in its early stages with numerous challenges awaiting resolutions, necessitating further explorations and developments.

In this survey, we present a comprehensive investigation of existing research on long video generation, aiming to provide a clear overview of the current state of development and contribute to its future progress. 
The organization of the rest of this paper is outlined in Figure 2. 
Initially,  we define the long video duration in Section 1. Section 2 discusses four different types of video generation models and control signals.
According to section 1 and section 2, we introduced two common paradigms to simplify long video generation tasks: \textit{Divide And Conquer} and \textit{Temporal AutoRegressive} in Section 3.1 and Section 3.2, respectively. 
Video quality improvement and hardware requirements are discussed in Section 4 and Section 5. Finally, the paper concludes with a summary of long video generation and a discussion on emerging trends and opportunities in Section 6.
\begin{figure*}[h]
    \centering
    \includegraphics[width=1\textwidth]{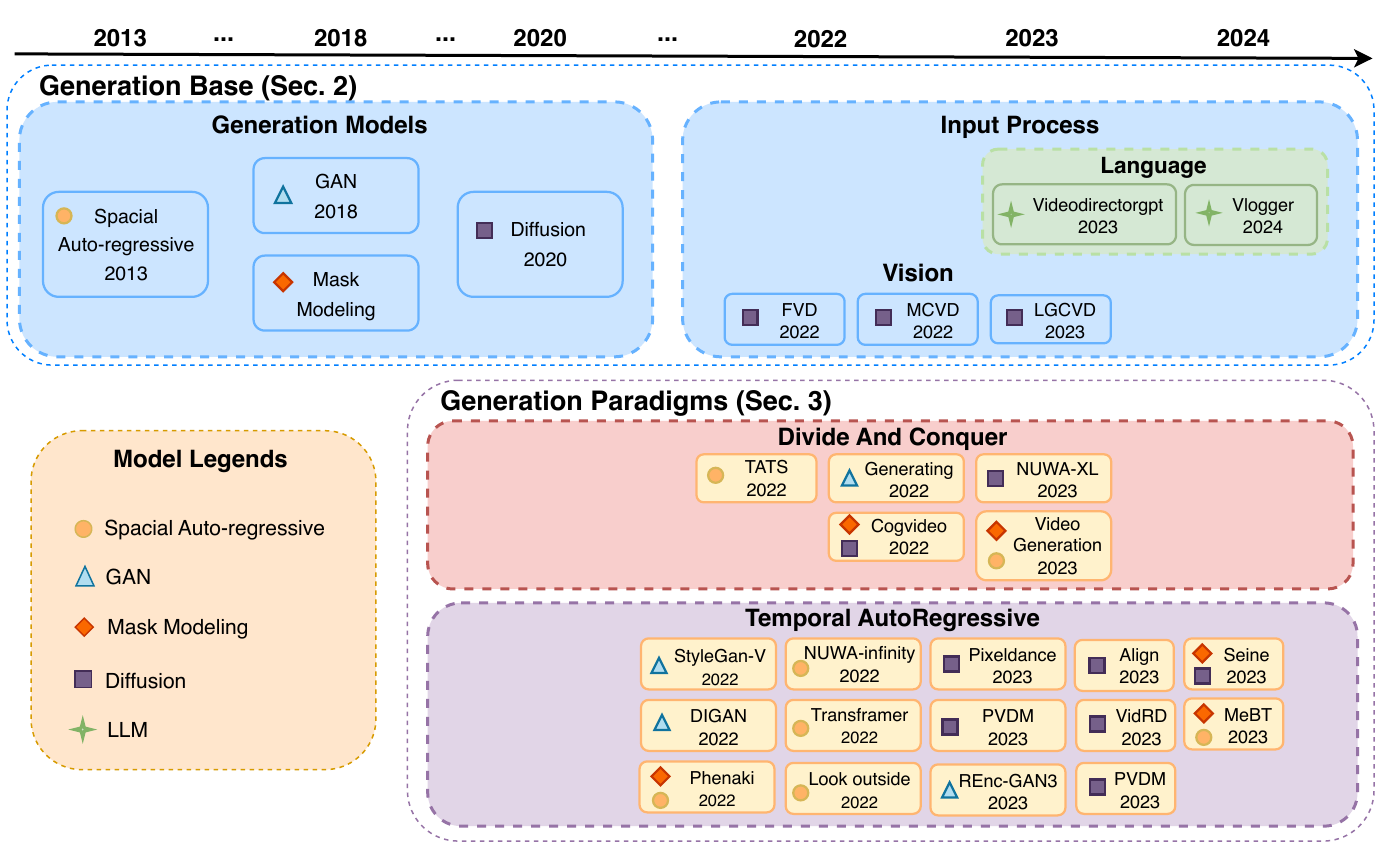}
    \caption{Evolution of Long Video Generation Techniques. This timeline provides an overview of foundational models and paradigms shaping long video generation, aligned with the in-depth discussions in subsequent sections.}
    \label{figure2}
\end{figure*}
\section{Basic Video Generation Techniques}
\label{sec: four video generation models}

\subsection{Four types of video generation models}
We detail the four types of popular video generation models, including Diffusion, Autoregressive, Generative Adversarial Networks (GAN), and Mask Modeling.

\noindent\textbf{Diffusion}
models for video generation adapt the iterative refinement process of traditional diffusion techniques, which were initially designed for static images \citep{DDPM}, to the dynamic realm of videos. At their core, these models start with a sequence of random noise and progressively denoise it through a series of steps to generate a coherent video sequence. Each step is guided by learned gradients that predictively denoise based on both the spatial content of individual frames and the temporal relationships between successive frames. This approach allows for the generation of videos where each frame is not only visually consistent with its predecessor but also contributes to the fluidity of the entire sequence.

\noindent\textbf{Spatial Auto-regressive} 
In video generation, spatial auto-regressive models\citep{auto} take a unique approach by synthesizing content through a patch-based methodology, where each patch's creation is dependent on the spatial relationship with previously generated patches. This process mirrors a recursive algorithm, generating one patch at a time. Thereby, it constructs the video frame-by-frame until completion. Within this framework, the spatial relationships between patches are crucial, as each subsequent patch must align seamlessly with its neighbors to ensure visual coherence across the entire frame. This method leverages the spatial dependencies inherent in video content, ensuring that as the video progresses temporally, each frame remains consistent and continuous with its predecessor, not just in time but also in space.

\noindent\textbf{GAN} (Generative Adversarial Networks) \citep{GAN}
In video generation using GAN, the process starts with the generator that turns a simple noise pattern into a sequence of video frames. This noise, essentially random, serves as the initial blank state from which the video is crafted. Through layers of neural networks, the generator progressively shapes this noise into images that look like frames of a video, ensuring each frame logically follows the last to create smooth motion and a believable narrative.

This evolution from noise to video is refined by feedback from the discriminator, a component that judges if the generated video looks real or fake. The generator learns from this judgment, improving its ability to produce more realistic videos over time. The ultimate goal is to generate the video to be indistinguishable from the real footage and showcase natural movements and transitions.

\noindent\textbf{Mask Modeling}
In video generation, mask modeling leverages the concept of selectively obscuring parts of video frames to enhance the model's learning process. This technique starts by applying masks to certain segments of the video, effectively hiding them from the model during training. The model then learns to predict these masked parts based on the visible context and the temporal flow of the video. This process not only forces the model to understand the underlying structure and dynamics of the video content but also improves its ability to generate coherent and continuous video sequences. By iterative training on partially visible data, the model becomes adept at filling in missing information, ensuring that generated videos maintain a natural progression of scenes and actions.
\subsection{Control signals for video generation}
In the long video generation process, control signals are normally required to control the video contents and styles. This section elaborates on the utilization of various types of prompts as control signals, each offering unique advantages in the generation process.

\noindent\textbf{Text prompts} guide the model to generate relevant video content for the description by specifying aspects such as theme, plot, and character behavior through text descriptions. 

\noindent\textbf{Image prompts} influence the visual style, scenes, or objects within the generated videos. By referencing input images, the model can produce content that is visually coherent and contextually related to the prompts, thereby enhancing the aesthetic and thematic relevance of the video output.

\noindent\textbf{Video prompts} enable the maintenance of stylistic, action-oriented, and emotional continuity from the input video in the generated content. They open pathways for varied generation techniques, including extrapolation, interpolation, the linkage of multiple video prompts, and video-to-video generation methods. These techniques facilitate the production of seamless, continuous long videos.

In existing researches on long video generation, common prompt inputs can be either a single type of prompt or a combination of multiple prompts, such as text and image or text and video. These control signals are instrumental in determining the content, style, and thematic direction of long videos.
\section{Long Video Generation Paradigms}
In the realm of long video generation, the challenges of limited computational resources and the incapacity of existing models to directly produce videos of significant duration have led to the proposal of two distinct paradigms: Divide And Conquer and Temporal Autoregressive, as depicted in Figure 3. These paradigms are designed to deconstruct the complex task of long video generation into more manageable processes, focusing on creating individual frames or short clips that can be logically assembled to fulfill the generation of long video.

The Divide And Conquer paradigm begins by identifying keyframes that outline the main narrative, followed by generating the intervening frames to weave together a cohesive long video. On the other hand, the Temporal Autoregressive paradigm, also known simply as Autoregressive, adopts a sequential approach to generate short video segments based on prior conditions. This paradigm aims to ensure a fluid transition between clips, thereby achieving a continuous long video narrative. Unlike Divide And Conquer, which adopts a hierarchical methodology by differentiating between storyline keyframes and supplementary filling frames, the Temporal Autoregressive paradigm forgoes a hierarchical structure, focusing instead on the direct generation of detailed clips informed by preceding frames.

In this section, the discussion focuses on both paradigms, examining how current research strategically simplifies the task of long video generation into smaller, more manageable tasks. Furthermore, it highlights how existing models are utilized for generation, with these outputs subsequently assembled to form a complete video narrative.
\begin{figure*}[h]
    \centering
    \includegraphics[scale=0.6]{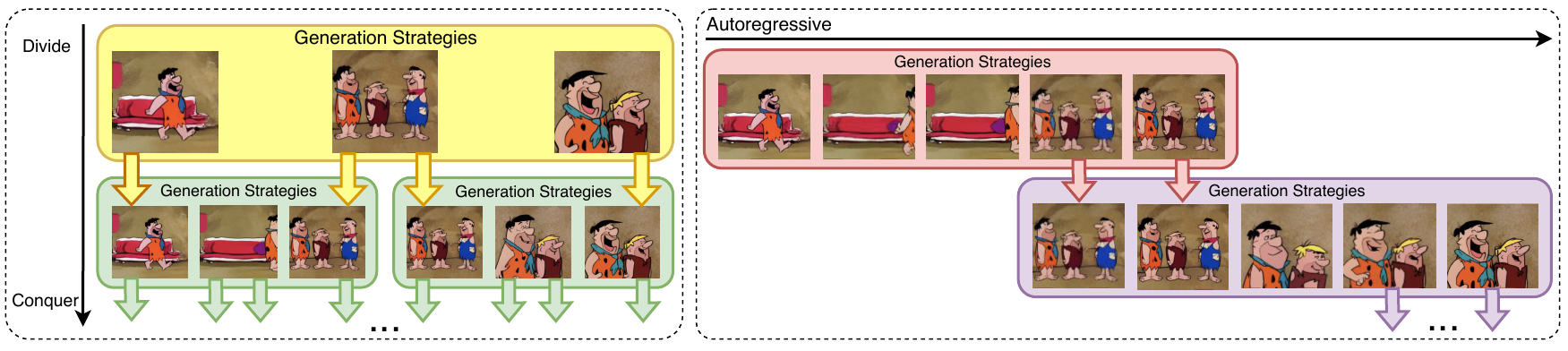}
    \caption{Overview of paradigms. The hierarchical generation process of Divide and Conquer is demonstrated by \citep{NUWA-XL}. For ease of understanding, generating long videos using the Autoregressive paradigm is demonstrated with identical video frames.}
    \label{figure3}
\end{figure*}
\subsection{Divide and Conquer}
\label{sec: divide and conquer}
The divide-and-conquer paradigm has emerged as a cornerstone in enhancing long video generation models. This approach bifurcates the complex process of generating long video sequences into manageable tasks, facilitating models to specialize in distinct aspects of video creation: keyframe generation and frame filling. We delve into how this paradigm has been tailored for long video generation as follows.

\noindent\textbf{Hierarchical Architecture for Frames Generation}
In the realm of long video generation, employing a divide-and-conquer methodology has led to the development of hierarchical architectures. These are designed to first generate keyframes that outline the video's storyline. Then, these focus on filling frames to complete the narrative. Global models excel in crafting storyline keyframes, while local models fill in the narrative gaps. In particular, \cite{NUWA-XL} showcased an architecture that leverages 3D-UNet-based diffusion models, aimed at this segmented approach. \cite{TATS} introduced a hierarchical transformer architecture which was designed to enhance temporal sensitivity and interpolation within long video narratives. 
This hierarchical structure allows for the incorporation of different models to fulfill distinct roles in the video creation process, collectively streamlining the production of consistent and seamless long video sequences.

\noindent\textbf{Staged Model Architecture for Resolution Enhancement}
To accommodate the flexibility of long videos, \cite{Generating} introduce a novel application of the divide-and-conquer strategy. They focus on initially generating low-resolution sequences, which are then enhanced to high resolution. This is achieved through a combination of a low-resolution GAN for coarse scene setting and a super-resolution StyleGAN3 for detail refinement. Such an approach effectively captures the essence of motion and narrative development across extended durations, reaching the goal of high-resolution long video generation. Furthermore, this application of the divide-and-conquer paradigm not only facilitates the production of long videos but also adapts to varying resolution needs, broadening the scope of long video generation to include enhanced detail and clarity.

\noindent\textbf{Streamlining Model Architecture through Mask Modeling}
In the context of long video generation, simplifying model architecture through mask modeling has been effective in merging the creation of keyframes and the process of filling frames into a unified and more streamlined procedure. \cite{Cogvideo} masked different conditions as inputs for keyframes and filling frames generation, which can simplify the Global and Local Diffusion models into a single model. \cite{huang2023video} focus on streamlining the generation of detailed scenes and narrative elements through masked layout integration, ensuring that key narrative points are effectively emphasized and interwoven throughout the long video.

\subsection{Temporal Autoregressive}
\label{sec: temporal autoregressive}
The Temporal Autoregressive generation paradigm offers another methodology within the domain of long video generation, drawing upon principles from time series analysis to streamline the generation process. This approach involves training the model on sequences of video data, where the output of one timestep serves as the input for the subsequent timestep. This iterative prediction mechanism facilitates the creation of videos that are not only long-term consistent but also coherently structured. Within this framework, the information from the previous timestep acts as the preceding condition, playing a pivotal role in guiding the model's predictions for future timesteps. Thus, it is critical to effectively manage these preceding conditions and accurately forecast subsequent frames, which contribute to the success of the Temporal Autoregressive paradigm and solid foundation for generating extended, cohesive video narratives.

In the following part, we divide the works under the Temporal Autoregressive paradigm into four categories according to their base generation model for generating videos. Each category employs unique methods to ensure smooth video progression, manage temporal-spatial details, and work efficiently. This structured division allows us to delve into how each model's architecture and methodology contribute significantly to achieving the objective of producing long, coherent, and visually appealing videos.
\subsection{Diffusion Models with AutoRegressive}
In the autoregressive paradigm, the process of long video generation is streamlined into sequentially creating video clips informed by preceding conditions, thereby easing the complexity of producing long videos. By utilizing this paradigm, diffusion models incorporate a latent space data representation to efficiently manage these preceding conditions. Subsequently, they refine the architecture of the model to boost the consistency of future predictions. This integration of the autoregressive paradigm with enhancements in diffusion models' design enables the production of seamless and temporally coherent long video narratives, advancing the video clip by clip, informed by a rich context of accumulated data.

\noindent\textbf{Compression for Preceding Conditions}
To adeptly handle the intricate data attributes of video content while optimizing computational and storage resources, recent research on diffusion models have explored a variety of compression strategies. Techniques range from condensing video data into a unified 3D latent space \citep{pixeldance,VidRD}, preserving critical features across dimensions and innovative approaches that \cite{PVDM} segregate and compress time and spatial information into distinct 2D spaces. This diversification in compression methods aims to balance the fidelity of feature representation with computational efficiency.

\noindent\textbf{Adding Temporal Layers for Long Video Clips Generation} 
The task of long video generation has been refined to focus on producing individual video clips, prompting adjustments to model architectures for improved output quality. Key developments include integrating temporal layers—such as attention and convolutional layers—into diffusion models \citep{align,VidRD}, enabling them to grasp the intricacies of video temporality. This enhancement transforms latent diffusion models into capable video generators by leveraging preceding conditions and facilitating the iterative creation of video clips, paving the way for long video generation.

\noindent\textbf{Refined Training and Generation Strategies}
Beyond model architecture, the adoption of sophisticated training methodologies significantly boosts the model's ability to replicate long video sequences. This includes dual-phase training that encompasses both unconditional learning of data distributions and conditional generation based on specific preceding conditions \citep{PVDM}. Furthermore, the introduction of a reuse strategy, which iteratively applies and removes noise to simulate the natural variability inherent in video content, represents a novel approach to enhancing model performance \citep{VidRD}.
\subsection{Spatial Auto-regressive Models with Temporal AutoRegressive}
In this section, we explore how spatial autoregressive models, particularly those transformer architectures, are adapted to generate long videos within the temporal autoregressive paradigm. These models are particularly adept at processing tokenized sequence-style inputs, enabling the segmentation of video samples into patches for more efficient processing and modeling. To further augment their capacity for capturing the essence of long videos, research has focused on enhancing the models' scalability and modeling capabilities. The fusion of video data features with the autoregressive capabilities of transformer models significantly enhances their ability to generate long video sequences.

\noindent\textbf{Tokenize Preceding Conditions}
For efficient data handling, researchers like \cite{NUWA-infinity} transform video frames into variable-sized patches, incorporating positional data for spatial context. Similarly, to conserve computational resources, both autoregressive transformer models and diffusion models compress video data into a latent space. Techniques vary from using sparse Discrete Cosine Transform (DCT) features that highlight interframe differences \citep{transframer} to employing VQGAN for converting frames into latent tokens \citep{Look}, optimizing both spatial and temporal data representation.

\noindent\textbf{Scaling Modeling Capabilities for Long Video Generation}
To address the transformer's inherent limitations with long-range temporal and large-scale spatial information, recent advancements introduce architectural modifications. By incorporating specialized blocks, such as attention mechanisms tailored for spatiotemporal data, these models can better manage the complexities of long video generation. For instance, \cite{transframer} integrate temporal and spatial annotations through cross-attention mechanisms, enhancing the model's predictive accuracy for sequential video frames. Similarly, \cite{Look} leverage self-attention to meld camera motion trajectories with frame data, enabling the autoregressive transformer to accurately forecast new camera positions and their associated imagery.
\subsection{GAN Models with Temporal AutoRegressive}
In this section, we delve into the utilization of GAN models within the temporal autoregressive paradigm for long video generation. GAN models, renowned for their generator-discriminator architecture, innovatively dissect preceding conditions into two fundamental elements: context and motion. This division allows for a more refined approach to spatiotemporal modeling in long video generation, leveraging dynamic-static separation techniques to distinguish between moving elements and static backgrounds in video sequences.

\noindent\textbf{Dynamic-Static Separation for Preceding Conditions}
By dividing video samples into contextual and movement aspects, GANs enhance their ability to faithfully reproduce both the evolving actions and the invariant scenes within videos \citep{REnc-GAN3}. Advanced methodologies, including neural network encoding as suggested by \cite{Stylegan-v} and \cite{DIGAN}, convert video inputs into Information Representations (INRs). These representations, comprising content and motion vectors, encapsulate the spatial and temporal attributes essential for detailed video sequence modeling \citep{DIGAN}.

\noindent\textbf{Innovations in Generator Design}
Common convolutional networks, coupled with strategic noise introduction, form the basis of new frame generation in latent space \citep{Stylegan-v, DIGAN}. With the incorporation of autoregressive principles, \cite{REnc-GAN3} propose a recall mechanism that ensures a seamless temporal linkage between video clips, marked by a half-frame overlap for uninterrupted motion flow.

\noindent\textbf{Advancements in Discriminator Design}
This section, moving beyond the conventional dual-stream discriminators \citep{REnc-GAN3}, delves into innovative approaches that enhance the discriminative capabilities of GAN models for long video sequences. Notably, \cite{Stylegan-v} have developed a hyper network-based method that integrates image and video discrimination into a unified process, streamlining the evaluation of video content. Meanwhile, \cite{DIGAN} have focused on refining the evaluation techniques through relational analysis between frame pairs. This approach not only allows for a more nuanced discrimination of long video sequences but also contributes to resource efficiency. 
\subsection{Mask Modeling with Temporal AutoRegressive}
In this section, we delve into the integration of mask modeling with the temporal autoregressive paradigm, focusing on its synergistic application alongside diffusion models and bidirectional transformers. Mask modeling significantly enhances the adaptability in learning sample distributions during training and streamlines the reuse of models in the generation phase.

\noindent\textbf{Mask Modeling During Training} 
Innovations such as the use of probabilistic masks, based on the Bernoulli distribution or predetermined patterns to selectively obscure parts of input frames have been introduced by researchers including \cite{seine}, \cite{align}, \cite{Phenaki}, and \cite{MeBT}. This methodology enables the models to learn the underlying data distribution by comparing the loss function between unmasked and masked data.

\noindent\textbf{Mask Modeling During Generation} In the generation phase, mask modeling plays a crucial role in determining the next frames to be produced. By utilizing video and text prompts as initial conditions, the method dynamically updates these prompts by masking frames that are temporally distant and focusing on those closer in the sequence. This approach allows for a continuous refresh of context, enabling the autoregressive model to produce extended sequences with enhanced coherence and relevance \citep{Phenaki,MeBT}.
\section{Towards Photo-realistic Long Video Generation}
\label{sec: evaluation}
\begin{table}[t]
\centering
\caption{Dataset in Long Video Generation. This table details the datasets frequently employed in the study of long video generation. It includes comprehensive information regarding the number of videos, the average duration of video segments, video styles, and resolutions available in each dataset}
\label{table1}
\renewcommand{\arraystretch}{1.4}
\scalebox{0.6}{
\begin{tabular}{l|cccc}
\hline
Dataset                      & Quantity   & Duration     & Style               & Resolution   \\ \hline
UCF-101\citep{UCF-101}       & 13320      & -          & Multiple categories & $256\times256$ \\
Sky Time-lapse\citep{Sky}    & 5000       & -          & Natural scenery     & $256\times256$ \\
BAIR\citep{BAIR}             & 44000      & -          & Real life           & $64\times64$   \\
WebVid10M\citep{Web}         & 10 million & 18 seconds & Multiple categories & 360P         \\
kinetics-600\citep{kinetics} & 500000     & 10 seconds & Human action        & $256\times256$ \\
Taichi HD\citep{Taichi}      & 280        & -          & Human action        & $256\times256$ \\
MSR-VTT\citep{MSR-VTT}       & 10000      & -          & Multiple categories & 240P         \\
KTH\citep{KTH}               & 600        & 4 seconds  & Human actions       & -            \\ \hline
\end{tabular}
}
\end{table}

\label{sec: evaluation}
\begin{table*}[t]
\centering
\caption{Evaluation Metrics in Long Video Generation. This table enumerates the evaluation metrics widely used in the field of long video generation, organized by their assessment categories (Image/Frame level indicators, Video level indicators), alongside detailed descriptions and the dimensions of discrepancy. 
}
\label{table2}
\renewcommand{\arraystretch}{1.6}
\scalebox{0.67}{
\begin{tabular}{l|c|l|c}
\hline
Evaluation Metrics      & Category                                                                                  & \multicolumn{1}{c|}{Description}                                                                                                                                                                                        & Discrepancy Dimension        \\ \hline
FID\citep{FID}          & \multirow{5}{*}{\begin{tabular}[c]{@{}c@{}}Image/Frame\\ level\\ indicators\end{tabular}} & \begin{tabular}[c]{@{}l@{}}Evaluating the quality of a generated video by comparing synthesized \\ video frames with real video frames.\end{tabular}                                              & Keyframes quality            \\ \cline{1-1} \cline{3-4} 
PSNR\citep{PSNR}        &                                                                                           & Evaluating the ratio between the signal and noise in video frames.                                                                                                                                & Frames Signal-to-noise ratio \\ \cline{1-1} \cline{3-4} 
SSIM\citep{SSIM}        &                                                                                           & \begin{tabular}[c]{@{}l@{}}Evaluating the structural similarity between original and generated images/frames,\\ taking into account brightness, contrast, and structural similarity.\end{tabular} & Frames structural similarity \\ \cline{1-1} \cline{3-4} 
CLIPSIM\citep{CLIP-SIM} &                                                                                           & Evaluating the correlation between images/frames and text.                                                                                                                                        & Image and text correlation   \\ \cline{1-1} \cline{3-4} 
LPIPS\citep{LPIPS}      &                                                                                           & Evaluating the perceptual similarity between generated video frames and real video frames.                                                                                                        & Frames perceptual similarity \\ \hline
FVD\citep{FVD}          & \multirow{3}{*}{\begin{tabular}[c]{@{}c@{}}Video\\ level\\ indicators\end{tabular}}       & \begin{tabular}[c]{@{}l@{}}Evaluating the similarity of the generated video distribution in feature space to the real\\ video distribution.\end{tabular}                                          & Feature distribution         \\ \cline{1-1} \cline{3-4} 
KVD\citep{KVD}          &                                                                                           & Evaluating the quality of generated videos by using Maximum Mean Discrepancy (MMD).                                                                                                               & Maximum Mean Discrepancy     \\ \cline{1-1} \cline{3-4} 
IS\citep{IS}            &                                                                                           & \begin{tabular}[c]{@{}l@{}}Calculating the Inception score of generated videos using features extracted by\\ 3D-ConvNets (C3D).\end{tabular}                                                      & Inception scores             \\ \hline
\end{tabular}
}
\end{table*}
In Section 3, with the help of two paradigms that simplify long video generation tasks within resource constraints, existing models are enabled to generate long videos step by step. Despite advancements, challenges like frame skipping, motion inconsistencies, and abrupt scene transitions still remain, hindering the pursuit of photo-realism in long video generation. Research efforts are currently being focused on enhancing data processing and structural optimization to directly address these issues. In the following parts, we highlight significant improvements across three key dimensions crucial for photo-realistic outputs: temporal-spatial consistency, content continuity, and diversity of long videos.
\subsection{Improve Temporal-Spatial Consistency}
Achieving temporal-spatial consistency is pivotal for generating high-quality long videos. This consistency ensures the seamless visual and temporal flow throughout the video, harmoniously blending various spatial elements and temporal sequences. The paradigms that deconstruct complex temporal and spatial dynamics into digestible segments for video clips are crucial for preserving this consistency. By effectively managing these segments, the approaches ensure that each clip contributes to a coherent overall narrative, upholding the temporal-spatial integrity of the entire video.

Existing research can be broadly categorized into two main aspects. The first aspect is that researchers aim to tailor the architecture of models to better capture and integrate temporal-spatial features within video clips. The second aspect is the application of both implicit and explicit methods to model preceding conditions play a significant role in bolstering video clips' temporal-spatial consistency.

\noindent\textbf{Model Structural Enhancements} 
Adding layers to models enhances the modeling of temporal-spatial features. \cite{FVD} proposed a combination method to add a temporal attention layer after the spatial attention layer. This approach enables spatial attention within each frame, as well as attention across different temporal frames for the same spatial position. \cite{MCVD} introduce a module called SPATIN within the U-Net's residual block to transmit temporal-spatial dynamics across the up-sampling and down-sampling modules, thereby facilitating the generation of new frames by implicitly modeling temporal-spatial dynamics.

Another approach involves adding modules. \cite{NUWA-infinity} incorporated the Nearby Context Pool mechanism into the hierarchical autoregressive generation model internal design. This mechanism dynamically selects nearby contextual token information using self-attention, enabling the model to learn spatial characteristics and ensure long-term consistency in the generated videos.

\noindent\textbf{Preceding Conditions Modeling} 
The preceding conditions encompass abundant input information that determines the generation. Therefore, it is of great importance to search extract, and model their spatio-temporal information, which can be divided into implicit and explicit components.

In an implicit manner, \cite{PVDM} perform a decoupled encoding of videos considering shared background and motion content, resulting in three image-like 2D latent representations. To further enhance temporal consistency, \cite{align} introduce a temporal discriminator constructed with 3D convolutions in the decoder part of the autoencoder. This temporal discriminator is utilized to fine-tune the generated video data. \cite{VidRD} combine the ideas from the aforementioned studies. The U-Net model is enhanced by introducing temporal convolution and temporal attention layers, which improve temporal feature extraction and capture the relationships between frames. A compression method called T-KLVAE was developed by combining spatial convolution, temporal convolution, and attention mechanisms \citep{NUWA-infinity}. This method compresses pixel-level images into a 3D latent space, where both temporal and spatial information are jointly encoded.

In addition, explicit approaches decompose video signals along the temporal dimension into motion optical flow and motion content. \cite{Control-A-Video} divide motion prior into pixel residual and optical flow components, then employ a two-stage explicit modeling approach to enhance video consistency and continuity through noise handling strategies. \cite{MV-Diffusion} explicitly model global trajectories and local trends. It utilizes optical flow from past and future frames to explicitly model global motion trajectories. \cite{VideoFusion} decompose video frames and noise into a shared base component (base) and a time-varying residual component (residual), while the diffusion process is correspondingly decomposed into two parts. This method effectively reduces content redundancy between frames and enhances temporal correlation. 
\subsection{Improve Content Continuity}
Ensuring content continuity, in tandem with temporal-spatial consistency is vital for preserving the coherence of actions and narratives across long videos. To achieve this, it requires a seamless fusion of video clips and frames, underpinning the fluidity and natural progression of the video's storyline. The forthcoming section will delve into methodologies aiming at reinforcing continuity.

\noindent\textbf{Model Structural Enhancements } \cite{VideoFusion} decompose video frames and noise into a shared component and a residual component that varies along the temporal axis. This approach better captures changing features, reduces attention to irrelevant features, and avoids generating redundant content, thereby eliminating frame stuttering and motion artifacts.

\noindent\textbf{Training and Generating Strategies} 
\cite{MCVD} adopt a simpler approach by training directly on long videos to eliminate the gap between predicted and real long videos, thus achieving the objectives of continuity and consistency. \cite{VideoPoet} autoregressively generate new video segments rely on the last 1 second of existing videos as a condition. It can accurately capture the temporal features of the video and achieve natural and smooth transitions, ensuring coherence and consistency in the long video generation.
\subsection{Improve Diversity of Long Video}
Diversity in long video generation represents a critical area of exploration, setting long videos apart from the often homogeneous nature of short video content. To elevate the creativity and innovation within long videos, existing research has ventured into several key areas: varying size, improving resolution, introducing content elements, enriching the diversity of motions, and incorporating changes in views.

\noindent\textbf{Resolution Improvement and Varable Size}
To achieve the generation of high-resolution long videos, \cite{align} refine the high resolution to high spatial and temporal resolutions. During training, it uses video datasets with different temporal resolutions and introduces a mask modeling to mask the filling frames, thereby achieving the goal of generating high-frame-rate long videos. To achieve the generation of high-resolution and variable-size long videos, \cite{NUWA-infinity} employ a hierarchical framework to refine the video generation process into smaller patches and token generation. In the generation direction, a self-regressive strategy is employed based on the designed modules, allowing flexible extension of the patch count in both length and width dimensions.

\noindent\textbf{Changes of Views}
\cite{transframer} and \cite{consistent} focus on generating new views in long videos. They incorporate visual conditions from different views as part of the input and employ the model to enhance the structural modeling of views by adding layers, such as the epipolar attention layer. This enables the model to capture view information and facilitate generation.
\section{Computational, Memory, and Data Resources}
\label{sec: computation}
In earlier discussions, we addressed challenges and innovative methodologies in long video generation. However, the extended lengths of long videos, their complex feature interdependencies, and increased dimensionality intensify the demands of significant computational and memory resources.

To counter these resource constraints, recent studies have shifted focus toward developing more efficient models and smarter training strategies. Efforts to enhance the accessibility of high-quality long video datasets have also been initiated, aiming to reduce manual effort and time investment, thereby making it easier to obtain diverse and rich training materials. This section further details the datasets and evaluation metrics employed in current research, shedding light on the standards used to evaluate long video generation techniques.

\noindent\textbf{Resource-Saving Data Compression}
Techniques that compress data into latent spaces, thereby minimizing data dimensionality and eradicating redundant information, have proven to be effective. This results in a more compact data representation and reduced computational complexity. The T-KLVAE \citep{LGCVD} and VAE encoders \citep{VidRD,NUWA-XL} have been notable for compressing pixel-level and temporal information into a 3D latent space. 

For a more refined approach, certain studies have transitioned to compressing data into a 2D latent space, prioritizing spatial information. To adequately represent temporal dynamics, these methods introduce additional temporal attention and specially designed spatial layers into the decoder \citep{align}. Furthermore, \cite{PVDM} have utilized an autoencoder. They integrated a parameterized, shared temporal component into the latent variables. This approach optimizes the balance between data compression and preserving essential video information.

\noindent\textbf{Lightweight Model Design}
Attention has turned towards mask modeling methods, which utilize probabilistic binary masking aligned with specific task objectives to enhance a model's versatility across various generation tasks while minimizing the need for multiple models \citep{PVDM,seine,RaMViD}. This approach significantly reduces computational demands. The bidirectional transformer model has also been highlighted for enabling more efficient parallel data processing, thus lowering computational load and enhancing processing speed \citep{MeBT}.

\noindent\textbf{Resource-Saving Training Strategies}
To address the lack of extensive long video datasets and the high costs of training, some studies pre-train models on large-scale text-to-image datasets before fine-tuning with video datasets, which facilitates a smooth transition from image to long video generation with minimal reliance on extensive video data \citep{align}. It achieves the transfer from image generation to long video generation with minimal usage of video datasets.

\noindent\textbf{Supplementary Dataset Methods}
Beyond the training strategies designed to optimize data utilization, recent efforts have focused on mitigating the inherent scarcity of video datasets. A novel approach introduced by \cite{VidRD} tackles this challenge head-on by devising a methodology for decomposing, transforming, and regenerating video datasets from pre-existing sources. This process involves segmenting videos by action and augmenting them with text descriptions through the use of large language models, thereby enhancing the textual diversity associated with each video segment. Additionally, techniques such as random scaling and translation are applied to transform still images into pseudo-videos. This inventive method not only augments the visual richness of the dataset but also directly addresses the fundamental issue of dataset scarcity, paving the way for more diverse and comprehensive video generation research.

\noindent\textbf{Summary of Dataset and Evaluation Metrics} 

In Tables 1 and 2, we compile the video datasets and evaluation metrics widely used in the field of research on long video generation. These tables meticulously detail each dataset's video count, lengths, styles, and resolutions. In the realm of evaluation metrics, we incorporate assessments of human performance. We acknowledge that the nuanced and detailed aspects of these evaluations cannot be easily summarized in tables. Therefore, the tables thoughtfully outline the dimensions of the discrepancy, alongside categories and descriptions. These tables are crafted to serve as an informative reference for those delving into long video generation studies.
\section{Conclusion and Future Directions}
This paper provides a comprehensive review of the latest research advances in the field of long video generation. We systematically review four video generation models and delve into the paradigms for generating long videos based on these models, categorizing them into two main types: Divide and Conquer and AutoRegressive. Furthermore, our work includes a comprehensive summary of quality characteristics in long video generation. Detailed explanations are provided for existing research that aims to enhance these qualities. Research focusing on resource requirement solutions is also discussed. To further advance the field, we identify several promising directions for future development.

\noindent\textbf{Expansion of Data Resources}

\noindent Existing approaches face challenges in training long video generation models due to inadequate resources in long video datasets, which fail to meet the requirement of obtaining optimal model parameters through training data. Consequently, this leads to issues such as incoherent long video generation and content repetition. To address this concern, \cite{VidRD} propose a method that uses large language models and transforms existing video content to expand the datasets, effectively resolving the data scarcity problem. Further research can explore more efficient approaches to enrich long video datasets.

\noindent\textbf{Developing Unified Generation Approaches}

\noindent The existing paradigms for long video generation are summarized into two categories: divide and conquer and autoregressive. Although they are capable of generating long videos with existing models, each has its drawbacks. Specifically, divide and conquer suffers from a scarcity of long video training datasets, requires significant time for generation, faces challenges in predicting keyframes over long durations, and the quality of keyframes significantly affects the quality of the filling frames. AutoRegressive tends to accumulate errors and suffers from content degradation after multiple inferences. Overall, each paradigm has its strengths and weaknesses. Future research could aim to develop a high-quality unified paradigm that integrates the strengths of both paradigms to address their respective limitations.

\noindent\textbf{Generation with Flexible Length and Aspect Ratio}

\noindent Current research predominantly focuses on training and creating long video content with predetermined dimensions. However, the growing need for diverse video content and simulations of the real world necessitates the generation of videos with variable lengths and aspect ratios. Sora \citep{Sora} and FiT \citep{FiT} have made strides in this area, with Sora enabling flexible video size generation and FiT showcasing adaptability in both dimensions for image generation. Future studies will likely emphasize improving video generation's flexibility, aiming to enhance generative models' applicability in real-world settings and spur further innovation in video content utilization.

\noindent\textbf{Generation of Super-Long Videos}

\noindent In the survey depicted in Figure 1, the longest duration of long videos in existing research is 1 hour \citep{Stylegan-v}. However, in real life like cinema and driving simulations, video durations typically are 90 minutes or even longer. We designate these as "super-long videos". Hence, future research could focus on the generation of super-long videos and address challenges such as view transitions, character and scene developments, and enrichment of action and plot that arise with extended durations.

\noindent\textbf{Enhanced Controllability and Real-world Simulation}

\noindent In long video generation, current models operate like black boxes during the generation process and internally, making it challenging to understand the causes of errors, such as those violating physical laws (as demonstrated by Sora \citep{Sora}). Existing solutions lack insight into the problems' origins and intuitive, controllable remedies. Thus, there's a need for new methods and technologies to enhance our understanding and control over generation models, making them more suitable for real-world applications.

\appendix



\begin{thebibliography}{}

\bibitem[\protect\citeauthoryear{Bain \bgroup \em et al.\egroup }{2021}]{Web}
Max Bain, Arsha Nagrani, G{\"u}l Varol, and Andrew Zisserman.
\newblock Frozen in time: A joint video and image encoder for end-to-end retrieval.
\newblock In {\em Proceedings of the IEEE/CVF International Conference on Computer Vision}, pages 1728--1738, 2021.

\bibitem[\protect\citeauthoryear{Blattmann \bgroup \em et al.\egroup }{2023}]{align}
Andreas Blattmann, Robin Rombach, Huan Ling, Tim Dockhorn, Seung~Wook Kim, Sanja Fidler, and Karsten Kreis.
\newblock Align your latents: High-resolution video synthesis with latent diffusion models.
\newblock In {\em Proceedings of the IEEE/CVF Conference on Computer Vision and Pattern Recognition}, pages 22563--22575, 2023.

\bibitem[\protect\citeauthoryear{Brooks \bgroup \em et al.\egroup }{2022}]{Generating}
Tim Brooks, Janne Hellsten, Miika Aittala, Ting-Chun Wang, Timo Aila, Jaakko Lehtinen, Ming-Yu Liu, Alexei Efros, and Tero Karras.
\newblock Generating long videos of dynamic scenes.
\newblock {\em Advances in Neural Information Processing Systems}, 35:31769--31781, 2022.

\bibitem[\protect\citeauthoryear{Carreira \bgroup \em et al.\egroup }{2018}]{kinetics}
Joao Carreira, Eric Noland, Andras Banki-Horvath, Chloe Hillier, and Andrew Zisserman.
\newblock A short note about kinetics-600.
\newblock {\em arXiv preprint arXiv:1808.01340}, 2018.

\bibitem[\protect\citeauthoryear{Chen \bgroup \em et al.\egroup }{2023a}]{Control-A-Video}
Weifeng Chen, Jie Wu, Pan Xie, Hefeng Wu, Jiashi Li, Xin Xia, Xuefeng Xiao, and Liang Lin.
\newblock Control-a-video: Controllable text-to-video generation with diffusion models.
\newblock {\em arXiv preprint arXiv:2305.13840}, 2023.

\bibitem[\protect\citeauthoryear{Chen \bgroup \em et al.\egroup }{2023b}]{seine}
Xinyuan Chen, Yaohui Wang, Lingjun Zhang, Shaobin Zhuang, Xin Ma, Jiashuo Yu, Yali Wang, Dahua Lin, Yu~Qiao, and Ziwei Liu.
\newblock Seine: Short-to-long video diffusion model for generative transition and prediction.
\newblock {\em arXiv preprint arXiv:2310.20700}, 2023.

\bibitem[\protect\citeauthoryear{Deng \bgroup \em et al.\egroup }{2023}]{MV-Diffusion}
Zijun Deng, Xiangteng He, Yuxin Peng, Xiongwei Zhu, and Lele Cheng.
\newblock Mv-diffusion: Motion-aware video diffusion model.
\newblock In {\em Proceedings of the 31st ACM International Conference on Multimedia}, pages 7255--7263, 2023.

\bibitem[\protect\citeauthoryear{Ebert \bgroup \em et al.\egroup }{2017}]{BAIR}
Frederik Ebert, Chelsea Finn, Alex~X Lee, and Sergey Levine.
\newblock Self-supervised visual planning with temporal skip connections.
\newblock {\em CoRL}, 12:16, 2017.

\bibitem[\protect\citeauthoryear{Ge \bgroup \em et al.\egroup }{2022}]{TATS}
Songwei Ge, Thomas Hayes, Harry Yang, Xi~Yin, Guan Pang, David Jacobs, Jia-Bin Huang, and Devi Parikh.
\newblock Long video generation with time-agnostic vqgan and time-sensitive transformer.
\newblock In {\em European Conference on Computer Vision}, pages 102--118. Springer, 2022.

\bibitem[\protect\citeauthoryear{Gu \bgroup \em et al.\egroup }{2023}]{VidRD}
Jiaxi Gu, Shicong Wang, Haoyu Zhao, Tianyi Lu, Xing Zhang, Zuxuan Wu, Songcen Xu, Wei Zhang, Yu-Gang Jiang, and Hang Xu.
\newblock Reuse and diffuse: Iterative denoising for text-to-video generation.
\newblock {\em arXiv preprint arXiv:2309.03549}, 2023.

\bibitem[\protect\citeauthoryear{Harvey \bgroup \em et al.\egroup }{2022}]{FVD}
William Harvey, Saeid Naderiparizi, Vaden Masrani, Christian Weilbach, and Frank Wood.
\newblock Flexible diffusion modeling of long videos.
\newblock {\em Advances in Neural Information Processing Systems}, 35:27953--27965, 2022.

\bibitem[\protect\citeauthoryear{Heusel \bgroup \em et al.\egroup }{2017}]{FID}
Martin Heusel, Hubert Ramsauer, Thomas Unterthiner, Bernhard Nessler, and Sepp Hochreiter.
\newblock Gans trained by a two time-scale update rule converge to a local nash equilibrium.
\newblock {\em Advances in neural information processing systems}, 30, 2017.

\bibitem[\protect\citeauthoryear{Hong \bgroup \em et al.\egroup }{2022}]{Cogvideo}
Wenyi Hong, Ming Ding, Wendi Zheng, Xinghan Liu, and Jie Tang.
\newblock Cogvideo: Large-scale pretraining for text-to-video generation via transformers.
\newblock {\em arXiv preprint arXiv:2205.15868}, 2022.

\bibitem[\protect\citeauthoryear{H{\"o}ppe \bgroup \em et al.\egroup }{2022}]{RaMViD}
Tobias H{\"o}ppe, Arash Mehrjou, Stefan Bauer, Didrik Nielsen, and Andrea Dittadi.
\newblock Diffusion models for video prediction and infilling.
\newblock {\em arXiv preprint arXiv:2206.07696}, 2022.

\bibitem[\protect\citeauthoryear{Huang \bgroup \em et al.\egroup }{2023}]{huang2023video}
Hsin-Ping Huang, Yu-Chuan Su, and Ming-Hsuan Yang.
\newblock Video generation beyond a single clip.
\newblock {\em arXiv preprint arXiv:2304.07483}, 2023.

\bibitem[\protect\citeauthoryear{Huynh-Thu and Ghanbari}{2008}]{PSNR}
Quan Huynh-Thu and Mohammed Ghanbari.
\newblock Scope of validity of psnr in image/video quality assessment.
\newblock {\em Electronics letters}, 44(13):800--801, 2008.

\bibitem[\protect\citeauthoryear{Kondratyuk \bgroup \em et al.\egroup }{2023}]{VideoPoet}
Dan Kondratyuk, Lijun Yu, Xiuye Gu, Jos{\'e} Lezama, Jonathan Huang, Rachel Hornung, Hartwig Adam, Hassan Akbari, Yair Alon, Vighnesh Birodkar, et~al.
\newblock Videopoet: A large language model for zero-shot video generation.
\newblock {\em arXiv preprint arXiv:2312.14125}, 2023.

\bibitem[\protect\citeauthoryear{Liang \bgroup \em et al.\egroup }{2022}]{NUWA-infinity}
Jian Liang, Chenfei Wu, Xiaowei Hu, Zhe Gan, Jianfeng Wang, Lijuan Wang, Zicheng Liu, Yuejian Fang, and Nan Duan.
\newblock Nuwa-infinity: Autoregressive over autoregressive generation for infinite visual synthesis.
\newblock {\em Advances in Neural Information Processing Systems}, 35:15420--15432, 2022.

\bibitem[\protect\citeauthoryear{Alex Graves}{2013}]{auto}
Alex Graves.
\newblock Generating sequences with recurrent neural networks.
\newblock {\em arXiv preprint arXiv:1308.0850}, 2013.

\bibitem[\protect\citeauthoryear{Lin \bgroup \em et al.\egroup }{2023}]{videodirectorgpt}
Han Lin, Abhay Zala, Jaemin Cho, and Mohit Bansal.
\newblock Videodirectorgpt: Consistent multi-scene video generation via llm-guided planning.
\newblock {\em arXiv preprint arXiv:2309.15091}, 2023.

\bibitem[\protect\citeauthoryear{Luo \bgroup \em et al.\egroup }{2023}]{VideoFusion}
Zhengxiong Luo, Dayou Chen, Yingya Zhang, Yan Huang, Liang Wang, Yujun Shen, Deli Zhao, Jingren Zhou, and Tieniu Tan.
\newblock Videofusion: Decomposed diffusion models for high-quality video generation.
\newblock In {\em Proceedings of the IEEE/CVF Conference on Computer Vision and Pattern Recognition}, pages 10209--10218, 2023.

\bibitem[\protect\citeauthoryear{Nash \bgroup \em et al.\egroup }{2022}]{transframer}
Charlie Nash, Joao Carreira, Jacob Walker, Iain Barr, Andrew Jaegle, Mateusz Malinowski, and Peter Battaglia.
\newblock Transframer: Arbitrary frame prediction with generative models.
\newblock {\em arXiv preprint arXiv:2203.09494}, 2022.

\bibitem[\protect\citeauthoryear{Radford \bgroup \em et al.\egroup }{2021}]{CLIP-SIM}
Alec Radford, Jong~Wook Kim, Chris Hallacy, Aditya Ramesh, Gabriel Goh, Sandhini Agarwal, Girish Sastry, Amanda Askell, Pamela Mishkin, Jack Clark, et~al.
\newblock Learning transferable visual models from natural language supervision.
\newblock In {\em International conference on machine learning}, pages 8748--8763. PMLR, 2021.

\bibitem[\protect\citeauthoryear{Ren and Wang}{2022}]{Look}
Xuanchi Ren and Xiaolong Wang.
\newblock Look outside the room: Synthesizing a consistent long-term 3d scene video from a single image.
\newblock In {\em Proceedings of the IEEE/CVF Conference on Computer Vision and Pattern Recognition}, pages 3563--3573, 2022.

\bibitem[\protect\citeauthoryear{Salimans \bgroup \em et al.\egroup }{2016}]{IS}
Tim Salimans, Ian Goodfellow, Wojciech Zaremba, Vicki Cheung, Alec Radford, and Xi~Chen.
\newblock Improved techniques for training gans.
\newblock {\em Advances in neural information processing systems}, 29, 2016.

\bibitem[\protect\citeauthoryear{Schuldt \bgroup \em et al.\egroup }{2004}]{KTH}
Christian Schuldt, Ivan Laptev, and Barbara Caputo.
\newblock Recognizing human actions: a local svm approach.
\newblock In {\em Proceedings of the 17th International Conference on Pattern Recognition, 2004. ICPR 2004.}, volume~3, pages 32--36. IEEE, 2004.

\bibitem[\protect\citeauthoryear{Siarohin \bgroup \em et al.\egroup }{2019}]{Taichi}
Aliaksandr Siarohin, St{\'e}phane Lathuili{\`e}re, Sergey Tulyakov, Elisa Ricci, and Nicu Sebe.
\newblock First order motion model for image animation.
\newblock {\em Advances in neural information processing systems}, 32, 2019.

\bibitem[\protect\citeauthoryear{Skorokhodov \bgroup \em et al.\egroup }{2022}]{Stylegan-v}
Ivan Skorokhodov, Sergey Tulyakov, and Mohamed Elhoseiny.
\newblock Stylegan-v: A continuous video generator with the price, image quality and perks of stylegan2.
\newblock In {\em Proceedings of the IEEE/CVF Conference on Computer Vision and Pattern Recognition}, pages 3626--3636, 2022.

\bibitem[\protect\citeauthoryear{Soomro \bgroup \em et al.\egroup }{2012}]{UCF-101}
Khurram Soomro, Amir~Roshan Zamir, and Mubarak Shah.
\newblock A dataset of 101 human action classes from videos in the wild.
\newblock {\em Center for Research in Computer Vision}, 2(11), 2012.

\bibitem[\protect\citeauthoryear{Tseng \bgroup \em et al.\egroup }{2023}]{consistent}
Hung-Yu Tseng, Qinbo Li, Changil Kim, Suhib Alsisan, Jia-Bin Huang, and Johannes Kopf.
\newblock Consistent view synthesis with pose-guided diffusion models.
\newblock In {\em Proceedings of the IEEE/CVF Conference on Computer Vision and Pattern Recognition}, pages 16773--16783, 2023.

\bibitem[\protect\citeauthoryear{Unterthiner \bgroup \em et al.\egroup }{2018}]{KVD}
Thomas Unterthiner, Sjoerd Van~Steenkiste, Karol Kurach, Raphael Marinier, Marcin Michalski, and Sylvain Gelly.
\newblock Towards accurate generative models of video: A new metric \& challenges.
\newblock {\em arXiv preprint arXiv:1812.01717}, 2018.

\bibitem[\protect\citeauthoryear{Villegas \bgroup \em et al.\egroup }{2022}]{Phenaki}
Ruben Villegas, Mohammad Babaeizadeh, Pieter-Jan Kindermans, Hernan Moraldo, Han Zhang, Mohammad~Taghi Saffar, Santiago Castro, Julius Kunze, and Dumitru Erhan.
\newblock Phenaki: Variable length video generation from open domain textual description.
\newblock {\em arXiv preprint arXiv:2210.02399}, 2022.

\bibitem[\protect\citeauthoryear{Voleti \bgroup \em et al.\egroup }{2022}]{MCVD}
Vikram Voleti, Alexia Jolicoeur-Martineau, and Chris Pal.
\newblock Mcvd-masked conditional video diffusion for prediction, generation, and interpolation.
\newblock {\em Advances in Neural Information Processing Systems}, 35:23371--23385, 2022.

\bibitem[\protect\citeauthoryear{Wang \bgroup \em et al.\egroup }{2004}]{SSIM}
Zhou Wang, Alan~C Bovik, Hamid~R Sheikh, and Eero~P Simoncelli.
\newblock Image quality assessment: from error visibility to structural similarity.
\newblock {\em IEEE transactions on image processing}, 13(4):600--612, 2004.

\bibitem[\protect\citeauthoryear{Xiong \bgroup \em et al.\egroup }{2018}]{Sky}
Wei Xiong, Wenhan Luo, Lin Ma, Wei Liu, and Jiebo Luo.
\newblock Learning to generate time-lapse videos using multi-stage dynamic generative adversarial networks.
\newblock In {\em Proceedings of the IEEE Conference on Computer Vision and Pattern Recognition}, pages 2364--2373, 2018.

\bibitem[\protect\citeauthoryear{Xu \bgroup \em et al.\egroup }{2016}]{MSR-VTT}
Jun Xu, Tao Mei, Ting Yao, and Yong Rui.
\newblock Msr-vtt: A large video description dataset for bridging video and language.
\newblock In {\em Proceedings of the IEEE conference on computer vision and pattern recognition}, pages 5288--5296, 2016.

\bibitem[\protect\citeauthoryear{Yang and Bors}{2023}]{REnc-GAN3}
Jingbo Yang and Adrian~G Bors.
\newblock Enabling the encoder-empowered gan-based video generators for long video generation.
\newblock In {\em 2023 IEEE International Conference on Image Processing (ICIP)}, pages 1425--1429. IEEE, 2023.

\bibitem[\protect\citeauthoryear{Yang \bgroup \em et al.\egroup }{2023}]{LGCVD}
Siyuan Yang, Lu~Zhang, Yu~Liu, Zhizhuo Jiang, and You He.
\newblock Video diffusion models with local-global context guidance.
\newblock {\em arXiv preprint arXiv:2306.02562}, 2023.

\bibitem[\protect\citeauthoryear{OpenAI}{2024}]{Sora}
Openai.com.
\newblock Sora.
\newblock {\em Sora}, 2024.

\bibitem[\protect\citeauthoryear{Yin \bgroup \em et al.\egroup }{2023}]{NUWA-XL}
Shengming Yin, Chenfei Wu, Huan Yang, Jianfeng Wang, Xiaodong Wang, Minheng Ni, Zhengyuan Yang, Linjie Li, Shuguang Liu, Fan Yang, et~al.
\newblock Nuwa-xl: Diffusion over diffusion for extremely long video generation.
\newblock {\em arXiv preprint arXiv:2303.12346}, 2023.

\bibitem[\protect\citeauthoryear{Yoo \bgroup \em et al.\egroup }{2023}]{MeBT}
Jaehoon Yoo, Semin Kim, Doyup Lee, Chiheon Kim, and Seunghoon Hong.
\newblock Towards end-to-end generative modeling of long videos with memory-efficient bidirectional transformers.
\newblock In {\em Proceedings of the IEEE/CVF Conference on Computer Vision and Pattern Recognition}, pages 22888--22897, 2023.

\bibitem[\protect\citeauthoryear{Yu \bgroup \em et al.\egroup }{2022}]{DIGAN}
Sihyun Yu, Jihoon Tack, Sangwoo Mo, Hyunsu Kim, Junho Kim, Jung-Woo Ha, and Jinwoo Shin.
\newblock Generating videos with dynamics-aware implicit generative adversarial networks.
\newblock {\em arXiv preprint arXiv:2202.10571}, 2022.

\bibitem[\protect\citeauthoryear{Yu \bgroup \em et al.\egroup }{2023}]{PVDM}
Sihyun Yu, Kihyuk Sohn, Subin Kim, and Jinwoo Shin.
\newblock Video probabilistic diffusion models in projected latent space.
\newblock In {\em Proceedings of the IEEE/CVF Conference on Computer Vision and Pattern Recognition}, pages 18456--18466, 2023.

\bibitem[\protect\citeauthoryear{Zeng \bgroup \em et al.\egroup }{2023}]{pixeldance}
Yan Zeng, Guoqiang Wei, Jiani Zheng, Jiaxin Zou, Yang Wei, Yuchen Zhang, and Hang Li.
\newblock Make pixels dance: High-dynamic video generation.
\newblock {\em arXiv preprint arXiv:2311.10982}, 2023.

\bibitem[\protect\citeauthoryear{Zhang \bgroup \em et al.\egroup }{2018}]{LPIPS}
Richard Zhang, Phillip Isola, Alexei~A Efros, Eli Shechtman, and Oliver Wang.
\newblock The unreasonable effectiveness of deep features as a perceptual metric.
\newblock In {\em Proceedings of the IEEE conference on computer vision and pattern recognition}, pages 586--595, 2018.

\bibitem[\protect\citeauthoryear{Zhuang \bgroup \em et al.\egroup }{2024}]{vlogger}
Shaobin Zhuang, Kunchang Li, Xinyuan Chen, Yaohui Wang, Ziwei Liu, Yu~Qiao, and Yali Wang.
\newblock Vlogger: Make your dream a vlog.
\newblock {\em arXiv preprint arXiv:2401.09414}, 2024.

\bibitem[\protect\citeauthoryear{Ho \bgroup \em et al.\egroup }{2020}]{DDPM}
Ho, Jonathan and Jain, Ajay and Abbeel, Pieter.
\newblock Denoising diffusion probabilistic models.
\newblock {\em Advances in neural information processing systems}, pages 6840--6851, 2020.

\bibitem[\protect\citeauthoryear{Creswell \bgroup \em et al.\egroup }{2020}]{GAN}
Goodfellow, Ian and Pouget-Abadie, Jean and Mirza, Mehdi and Xu, Bing and Warde-Farley, David and Ozair, Sherjil and Courville, Aaron and Bengio, Yoshua.
\newblock Generative adversarial networks: An overview
\newblock {\em Communications of the ACM}, pages 139--144, 2020.

\bibitem[\protect\citeauthoryear{Lu \bgroup \em et al.\egroup }{2024}]{FiT}
Zeyu Lu and Zidong Wang and Di Huang and Chengyue Wu and Xihui Liu and Wanli Ouyangand Lei Bai.
\newblock FiT: Flexible Vision Transformer for Diffusion Model.
\newblock {\em arXiv preprint arXiv:2402.12376}, 2024.

\end{thebibliography}

\end{document}